\documentclass[a4paper]{jpconf}
\usepackage{cite}
\usepackage{graphicx}
\usepackage{multirow}
\usepackage[ruled,vlined]{algorithm2e}

\begin{document}
\title{Online detection of failures generated by storage simulator}

\author{Kenenbek Arzymatov, Mikhail Hushchyn, Andrey Sapronov, Vladislav Belavin, Leonid Gremyachikh, Maksim Karpov and Andrey Ustyuzhanin}

\address{National Research University Higher School of Economics}
\address{11 Pokrovsky Boulevard, Moscow, Russia, 109028}
\ead{karzymatov@hse.ru}

\begin{abstract}
Modern large-scale data-farms consist of hundreds of thousands of storage devices that span distributed infrastructure. Devices used in modern data centers (such as controllers, links, SSD- and HDD-disks) can fail due to hardware as well as software problems. Such failures or anomalies can be detected by monitoring the activity of components using machine learning techniques. In order to use these techniques, researchers need plenty of historical data of devices in normal and failure mode for training algorithms. In this work, we challenge two problems: 1) lack of storage data in the methods above by creating a simulator and 2) applying existing online algorithms that can faster detect a failure occurred in one of the components.

We created a Go-based (golang) package for simulating the behavior of modern storage infrastructure. The software is based on the discrete-event modeling paradigm and captures the structure and dynamics of high-level storage system building blocks. The package's flexible structure allows us to create a model of a real-world storage system with a configurable number of components. The primary area of interest is exploring the storage machine's behavior under stress testing or exploitation in the medium- or long-term for observing failures of its components.

To discover failures in the time series distribution generated by the simulator, we modified a change point detection algorithm that works in online mode. The goal of the change-point detection is to discover differences in time series distribution. This work describes an approach for failure detection in time series data based on direct density ratio estimation via binary classifiers. 
\end{abstract}

\section*{Introduction}

Disk-drive is one of the crucial elements of any computer and IT infrastructure. Disk failures have a high contributing factor to outages of the overall computing system. During the last decades, the storage system's reliability and modeling is an active area of research in industry and academia works~\cite{744151, 4292318, 816306}. Nowadays, the rough total amount of hard disk drives (HDD) and solid-state drives (SSD) deployed in data-farms and cloud systems passed tens of millions of units~\cite{7474362}. Consequently, the importance of early identifying defects leading to failures that can happen in the future can result in significant benefits. Such failures or anomalies can be detected by monitoring components' activity using machine learning techniques, named change point detection~\cite{aminikhanghahi2017survey, doi:10.1137/1.9781611972795.34, LIU201372}. To use these techniques, especially for anomaly detection, it is a necessity in historical data of devices in normal and failure mode for training algorithms. In this paper, due to the reasons mentioned above, we challenge two problems: 1) lack of storage data in the methods above by creating a simulator and 2) applying new online algorithms that can faster detect a failure occurred in one of the components~\cite{hushchyn2020online}. 

A Go-based (golang) package for simulating the behavior of modern storage infrastructure is created. The primary area of interest is exploring the storage machine's behavior under stress testing or exploitation in the medium- or long-term for observing failures of its components. The software is based on the discrete-event modeling paradigm and captures the structure and dynamics of high-level storage system building blocks. It represents the hybrid approach to modeling storage attached network~\cite{karpov, 10.7717/peerj-cs.271}. This method uses additional blocks with a neural network that tunes the internal model parameters while a simulation is running, described in~\cite{belavin}. This approach's critical advantage is a decreased requirement for detailed simulation and the number of modeled parameters of real-world system components and, as a result, a significant reduction in the intellectual cost of its development. The package's modular structure allows us to create a model of a real-word storage system with a configurable number of components. Compared to other techniques, parameter tuning does not require heavy-lifting changes within developing service~\cite{Mousavi_2017}. 

To discover failures in the time series distribution generated by the simulator, we modified a change point detection algorithm that works in online mode. The goal of the change-point detection is to discover differences in time series distribution. This work uses an approach for failure detection in time series data based on direct density ratio estimation via binary classifiers~\cite{hushchyn2020online}.

\section*{Simulator}

\subsection*{Internals}

\begin{figure}[h]
\includegraphics[width=20pc]{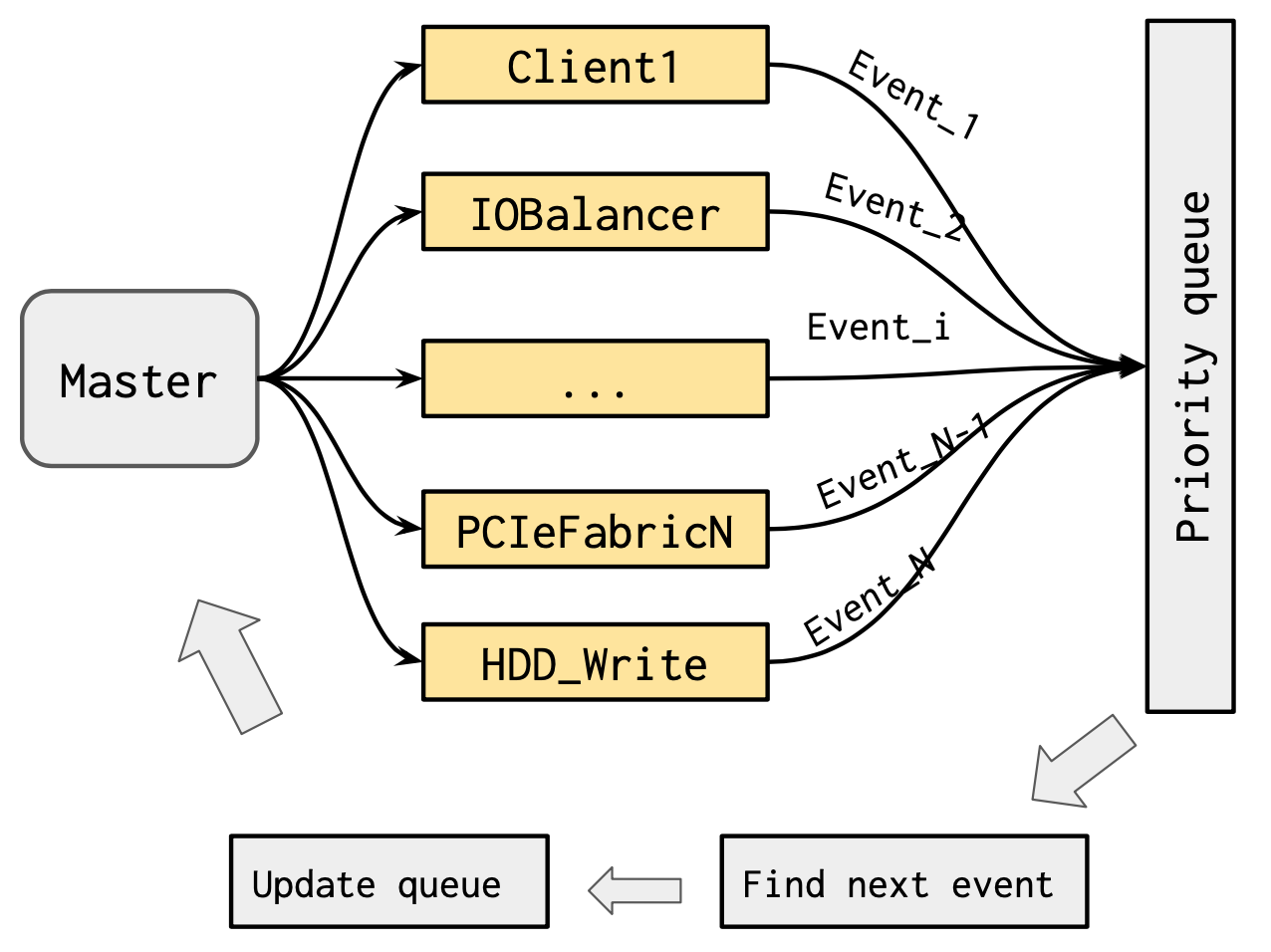}\hspace{2pc}%
\begin{minipage}[b]{14pc}\caption{\label{fig:loop} The event handling loop is the central part that responsible for time movement in the simulator. The Master process creates necessary logical processes (Client1, IOBalancer, HDD\_Write, etc.) and populates a Priority Queue by collecting events from modeling processes. The last part of the implementation is running the event handling loop. It removes successive elements from the queue. That would be correct because we know that the queue is already time sorted and performed the associated actions. }
\end{minipage}
\end{figure}

The simulator uses a Discrete Event Simulation (DES)~\cite{fishman} paradigm for modeling storage infrastructure. In a broad sense, DES is used to simulate a system as a discrete sequence of events in time. Each event happens in a specific moment in time and traces a change of state in the system. Between two consecutive events, no altering in the system is presumed to happen; thus, the simulation time can directly move to the next event's occurrence time. The scheme of the process is shown in Figure~\ref{fig:loop}. 

The simulator's programming environment provides the functionality to set up a model for specific computing environments, especially storage area networks. The key site of interest is exploring the storage infrastructure's behavior under various stress testing or utilization in the medium- or long-term for monitoring breakups of its components.

\begin{figure}[h]
\begin{minipage}{20pc}
\includegraphics[width=20pc]{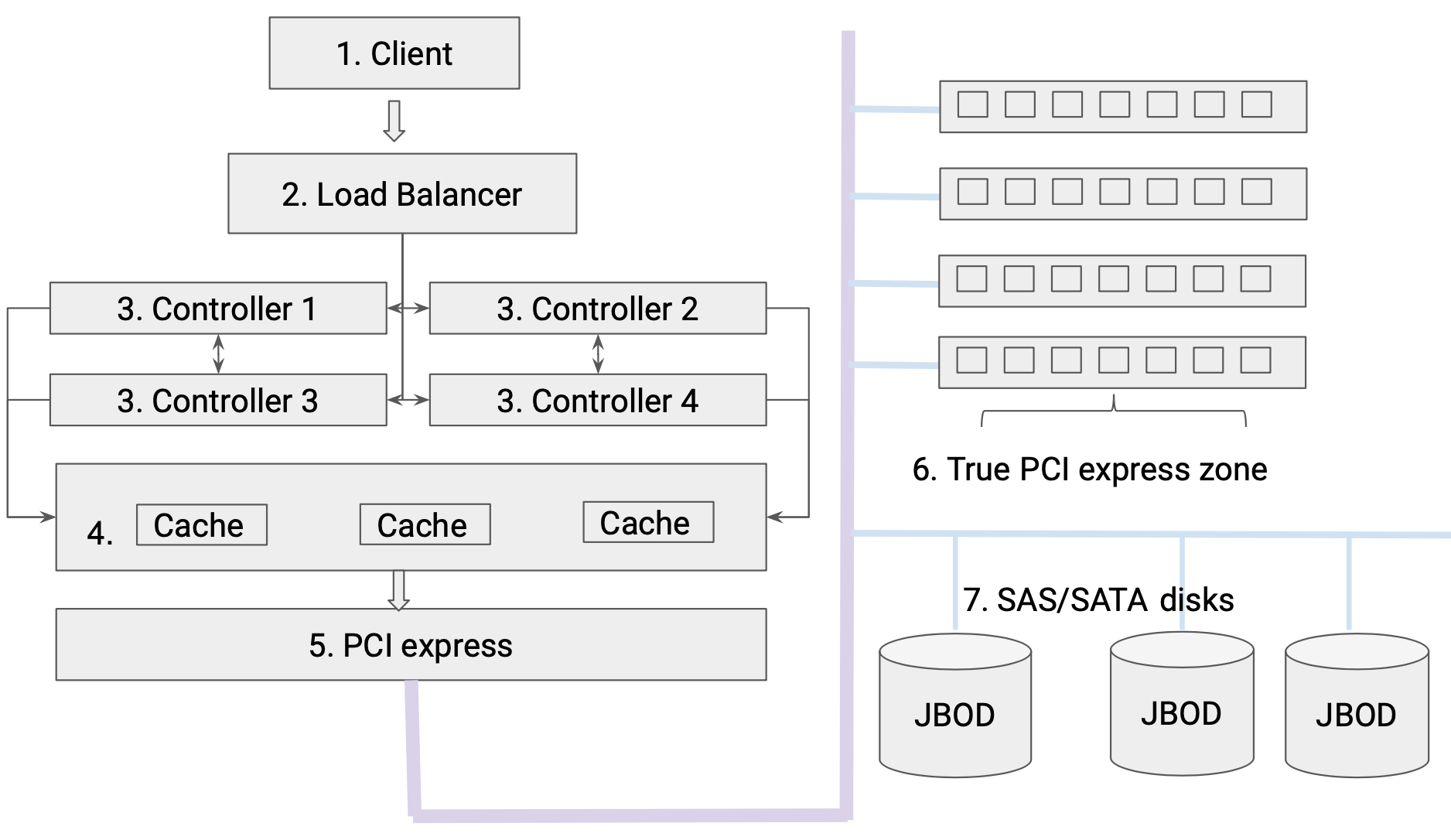}
\caption{\label{fig:realsystem}The example of the real storage system that can be modeled by using basic blocks}
\end{minipage}\hspace{1pc}%
\begin{minipage}{14pc}
\includegraphics[width=14pc]{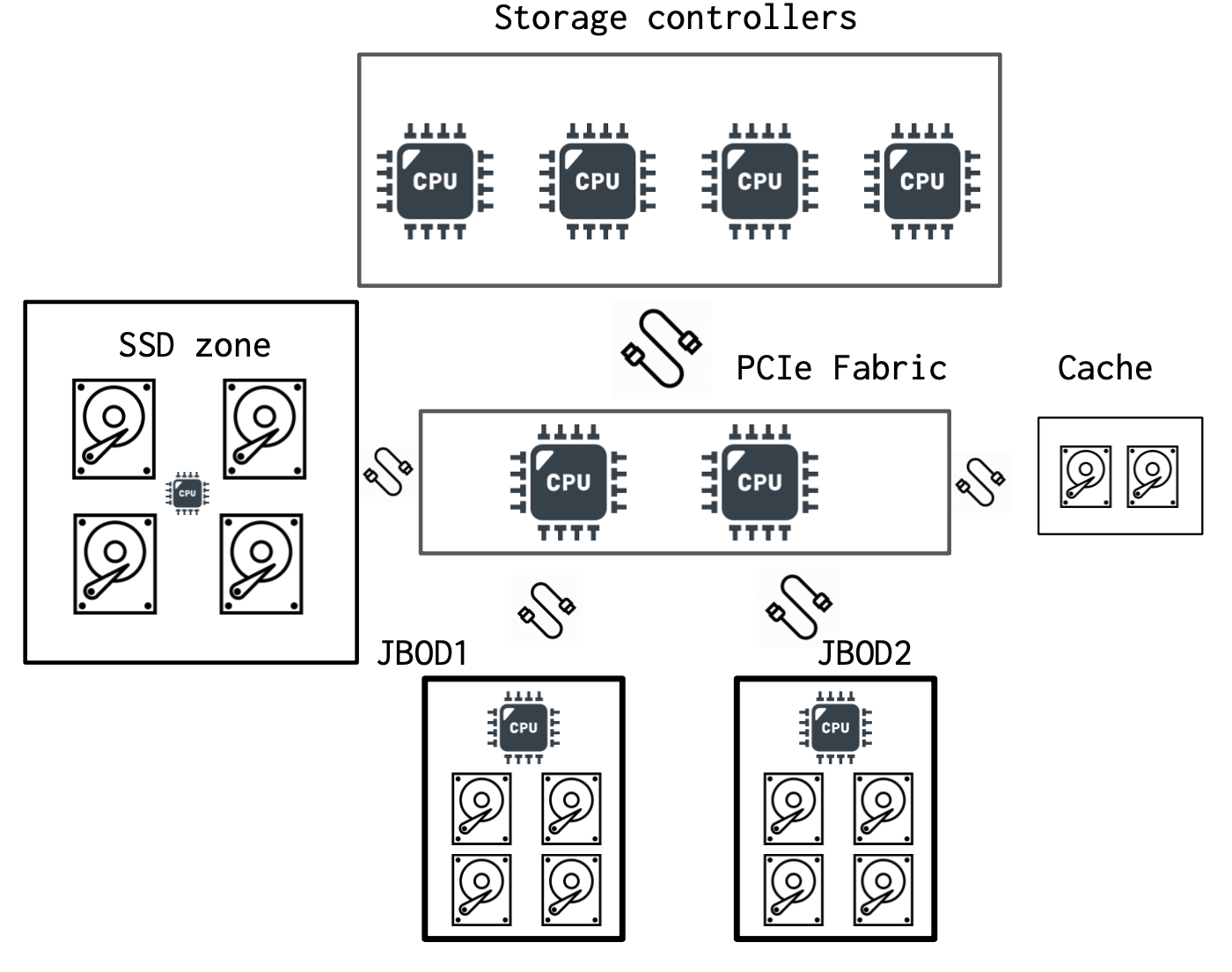}
\caption{\label{fig:basicblocks}Basic resource entities in the simulator}
\end{minipage} 
\end{figure}

In the simulator, load to storage system can be represented by two action types: read file from disk and write file to disk. Each file has corresponding attributes, such as name, block size, and total size. With the current load, these attributes determine the amount of time required to perform the corresponding action. The three basic types of resources are provided: CPU, network interface, and storage. Their representation is shown in the Figure~\ref{fig:basicblocks} and informative description is given in the Table~\ref{tab:res}. By using basic blocks, real-world systems can be constructed, as shown in the Figure~\ref{fig:realsystem}.

\begin{table} \caption{\label{tab:res} Resource description} \begin{center}
\begin{tabular}{lllll}
\br
Resource &Real word entity & Parameters & Units & Anomaly type \\
\mr
CPU & Controller, server & Number of cores & Amount  &  Each component \\
    & &  Core speed  & Flops & can suffer from  \\ 
Link & Networking cables &Bandwidth   & Megabyte/sec & performance \textbf{degradation} \\
     & & Latency    & Sec   & or total \textbf{breakup} \\
Storage & Cache, SSD, HDD & Size    & Gigabyte\\
        & & Write speed & Megabyte/sec \\
        & & Read speed  & Megabyte/sec  \\
\br
\end{tabular}
\end{center}
\end{table}

\subsection*{Comparison with the real data}

\begin{figure}[h]
    \begin{center}
    \includegraphics[width=0.7\textwidth]{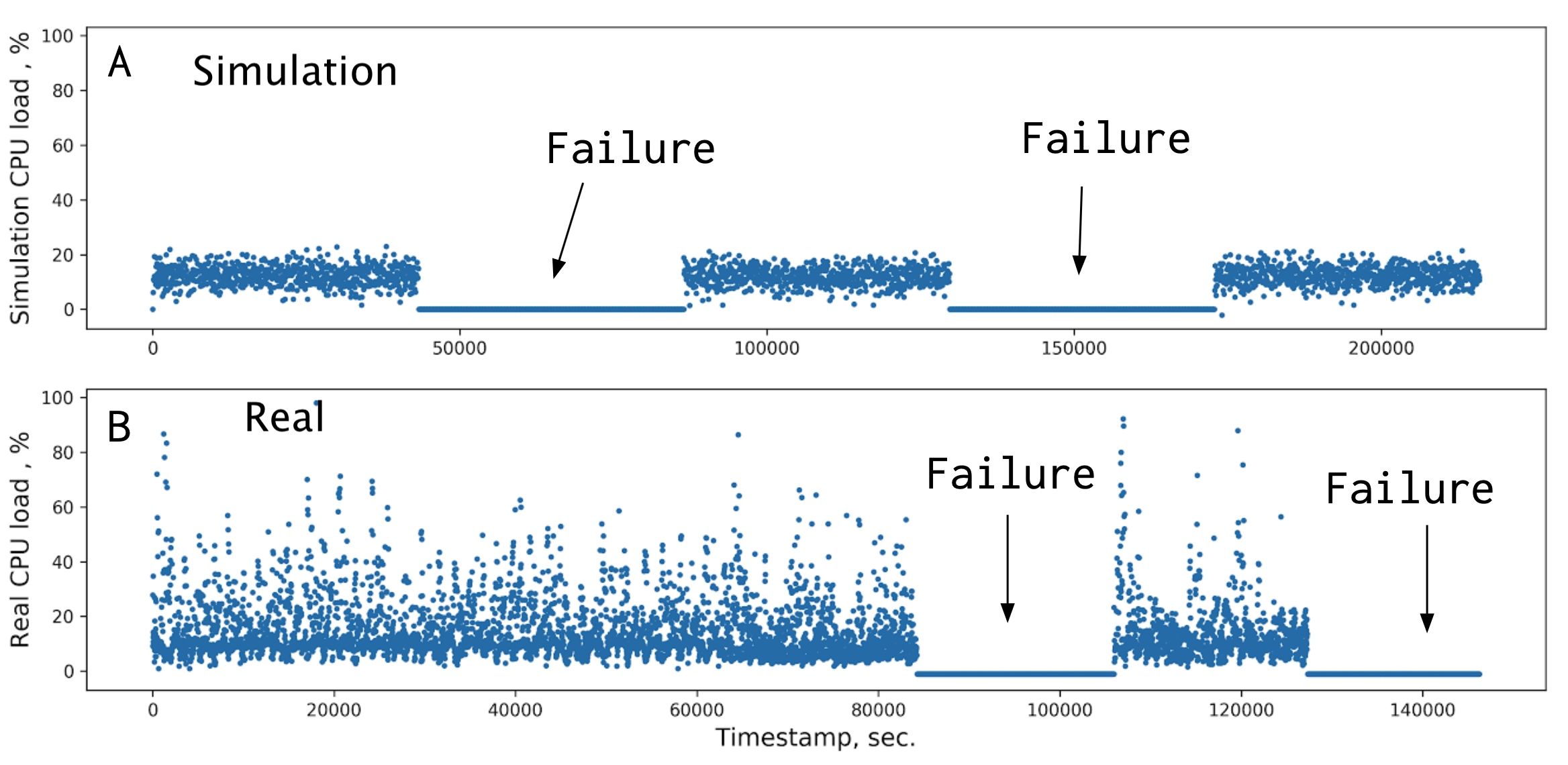} 
    \end{center}
    \caption{\label{fig:cmp}Comparison of the CPU load metrics between simulated (A) and real data (B). The periods marked ‘Failure’ correspond to a storage processor being offline}
\end{figure}

The data from the real-world storage system were used to validate the behavior of the simulator. A similar writing load scenario was generated on the model prototype, together with intentional controller failure (turn-off).  The comparison is shown in the Figure~\ref{fig:cmp}. As we can see, the simulator's data can qualitatively reflect the components breakup.

\section*{Change point detection}

    Consider a $d$-dimensional time series that is described by a vector of observations $x(t) \in R^d$ at time $t$. Sequence of observations for time $t$ with length $k$ is defined as:
    $$ X(t) = [x(t)^T, x(t-1)^T, \dots,x(t-k-1)^T]^T \in R^{kd} $$
    Sample of sequences of size $n$ is defined as:
    $$ \mathcal{X}(t) = {X(t), X(t-1), \dots, X(t - n + 1)} $$

It is implied that observation distribution changes at time $t^{*}$. The goal is to detect this change. The idea is to estimate dissimilarity score between reference $X_{rf}(t-n)$ and test $X_{te}(t)$. The larger dissimilarity, the more likely the change point occurs at time $t-n$. 
    
In this work, we apply a CPD algorithm based on direct density ratio estimation developed in~\cite{hushchyn2020online}. The main idea is to estimate density ratio $w(X)$ between two probability distributions $P_{te}(X)$ and $P_{rf}(X)$ which correspond to test and reference sets accordingly. For estimating $w(X)$, different binary classifiers can be used, like decision trees, random forests, SVM, etc. We use neural networks for this purpose. This network $f(X, \theta)$ is trained on the mini-batches with cross-entropy loss function $L(\mathcal{X}(t-l), \mathcal{X}(t), \theta)$,

$$
L(\mathcal{X}(t-l), \mathcal{X}(t), \theta) = - \frac{1}{n} \sum_{X \in \mathcal{X}(t-l)} \log (1 - f(X, \theta)) -  \frac{1}{n} \sum_{X \in \mathcal{X}(t)} \log f(X, \theta),
$$

We use a dissimilarity score based on the  Kullback-Leibler divergence, $D(\mathcal{X}(t-l), \mathcal{X}(t))$. Following~\cite{hushchyn2020generalization}, we define this score as:

$$    D(\mathcal{X}(t-l), \mathcal{X}(t), \theta) =  \frac{1}{n} \sum_{X \in \mathcal{X}(t-l)} \log \frac{1 - f(X, \theta)}{f(X, \theta)} +  \frac{1}{n} \sum_{X \in \mathcal{X}(t)} \log \frac{f(X, \theta)}{1 - f(X, \theta)}.
$$
%  $$ w(X) = \frac{P_{te}(X)}{P_{rf}(X)} =  \frac{P(X | y = 1)}{P(X | y=0)} $$
 
%  Due to Bayes rule we can rewrite $P(X | y=1)$ and $P(X | y=0)$, consequently $w(X)$ becomes:
 
% $$ w(X) = \frac{P(y=0)}{P(y=1)}\frac{P(y=1|X)}{P(y=0|X)} = \frac{P(y=1|X)}{P(y=0|X)} = \frac{f(x)}{1 - f(x)} $$

% As loss function, we use symmetric Kullback-Leibler divergence $D$ for calculating a distance between two distributions:

% $$ D = KL(P_{te} || P_{rf}) + KL(P_{rf} || P_{te}) $$
% $$ KL(P||Q) = \int P(x) log \frac{P(x)}{Q(x)}dx$$

% In a discrete form KL divergence transforms to: 

% $$ KL_{sym}(P||Q) = \frac{1}{n_{te}} \sum\limits_{X \in \mathcal{X}_{te}} \frac{f(X)}{1-f(X)} + \frac{1}{n_{rf}} \sum\limits_{X \in \mathcal{X}_{rf}} \frac{1-f(X)}{f(X)}  $$

According to~\cite{hushchyn2020online}, the training algorithm is shown in Alg.~\ref{alg:clf}. It consists of the following steps performing in the loop: 1) initializing hyper-parameters 2) preparing single datasets $\mathcal{X'}_{rf}$ and $\mathcal{X'}_{te}$ 3) calculating loss function $J$ 4) applying gradients to the weights of neural network.

\begin{algorithm}
\SetAlgoLined
\textbf{Inputs:} time series $\{X(t)\}_{t=k}^{T}$; $k$ -- size of a combined vector $X(t)$; $n$ -- size of a mini-batch $\mathcal{X}(t)$; $l$ -- lag size and $n \ll l$; $f(X, \theta)$ -- a neural network with weights $\theta$; \\
\textbf{Initialization:} $t \leftarrow k + n + l$; \\
\While{$t \le T$}{
 take mini-batches $\mathcal{X}(t-l)$ and $\mathcal{X}(t)$; \\
 $d(t) \leftarrow D(\mathcal{X}(t-l), \mathcal{X}(t), \theta)$; \\
 $\bar{d}(t) \leftarrow \bar{d}(t-n) + \frac{1}{l} (d(t) - d(t-l-n))$; \\
 $loss(t, \theta) \leftarrow L(\mathcal{X}(t-l), \mathcal{X}(t), \theta)$; \\
 $\theta \leftarrow \mathrm{Optimizer}(\mathrm{loss}(t, \theta))$; \\
 $t \leftarrow t + n$; \\
 }
 \Return{$\{\bar{d}(t)\}_{t=1}^{T}$} -- change-point detection score
 \caption{Change-point detection algorithm.}
 \label{alg:clf}
\end{algorithm}

% \begin{algorithm}[H]
% \SetAlgoLined
% \textbf{Inputs}: samples $\mathcal{X}_{rf}(t-n)$ and $\mathcal{X}_{te}(t)$, learning rate $\nu$, neural network weights $\theta$, number of epochs $M$, Adam hyperparameters $\nu$, $\beta_1$, $\beta_2$;
%  Initialize parameters $\theta_0$ of a neural network $\hat{w}(X, \theta_0)$;
 
% \textbf{for} $m=1, \dots, M $ do:
 
% \hspace*{20pt} for each batch do:
 
% \hspace*{40pt} $\mathcal{X'}_{rf}(t-n)$ $\leftarrow$
% \textbf{single example} from $\mathcal{X}_{rf}(t-n)$;
 
% \hspace*{40pt} $\mathcal{X'}_{te}(t)$ $\leftarrow$ \textbf{single example} from $\mathcal{X}_{te}(t)$; \newline

% \hspace*{40pt}$
% J(\theta) \leftarrow \frac{1}{n_{te}} \sum\limits_{X \in \mathcal{X}_{te}} \frac{\hat{w}(X)}{1-\hat{w}(X)} + \frac{1}{n_{rf}} \sum\limits_{X \in \mathcal{X}_{rf}} \frac{1-\hat{w}(X)}{\hat{w}(X)}$

% \hspace*{40pt}$ \theta \leftarrow Adam (\nabla_\theta J(\theta), \theta, \nu, \beta_1, \beta_2) $

% \textbf{return} Approximation function $\hat{w}(X, \theta_0)$

% \caption{Training algorithm}
% \end{algorithm}

\section*{Results}

\begin{figure}[h]
\begin{minipage}{19pc}
\includegraphics[width=19pc]{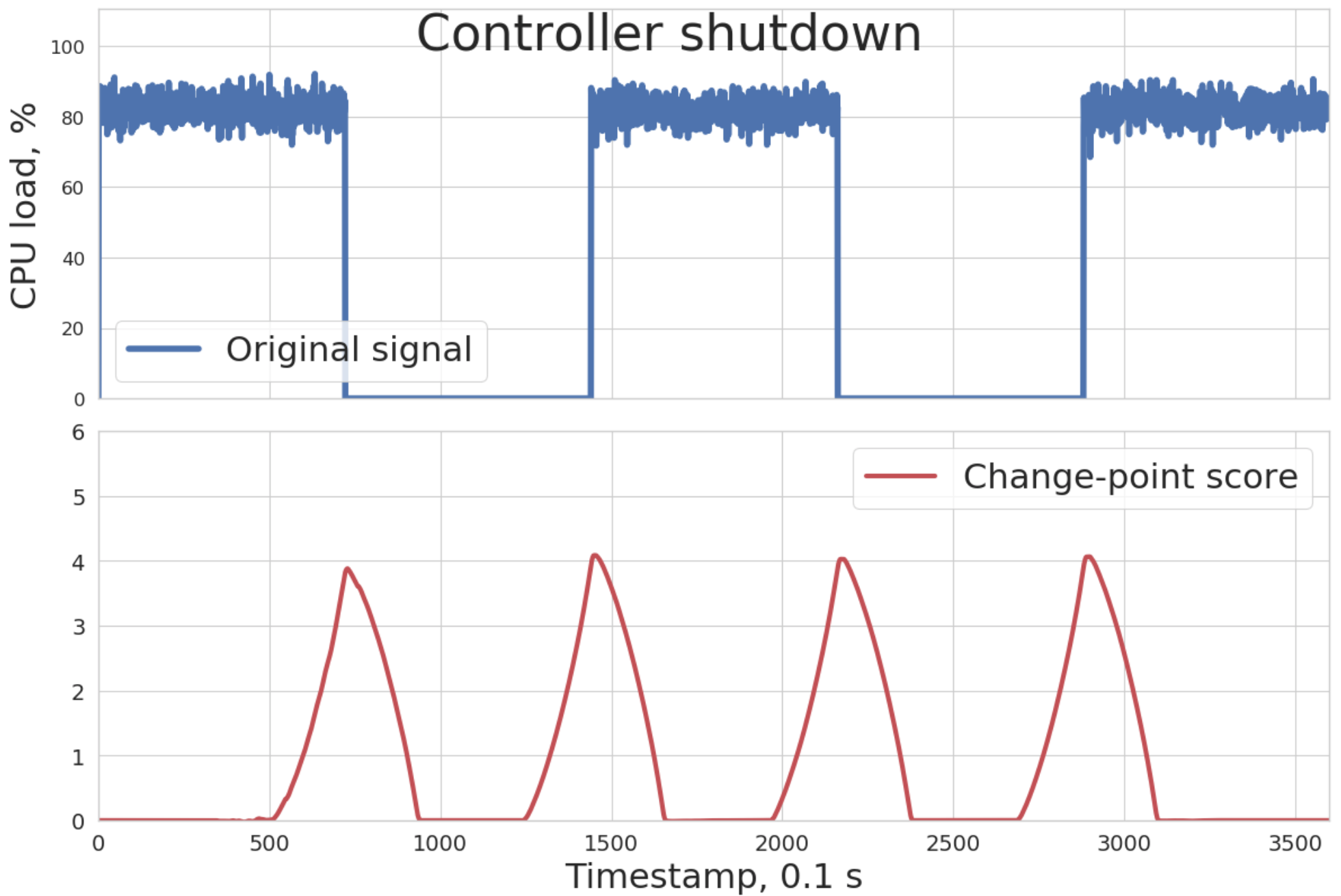}
\caption{\label{fig:cpd1}Controller failure}
\end{minipage}\hspace{1pc}%
\begin{minipage}{18pc}
\includegraphics[width=18pc]{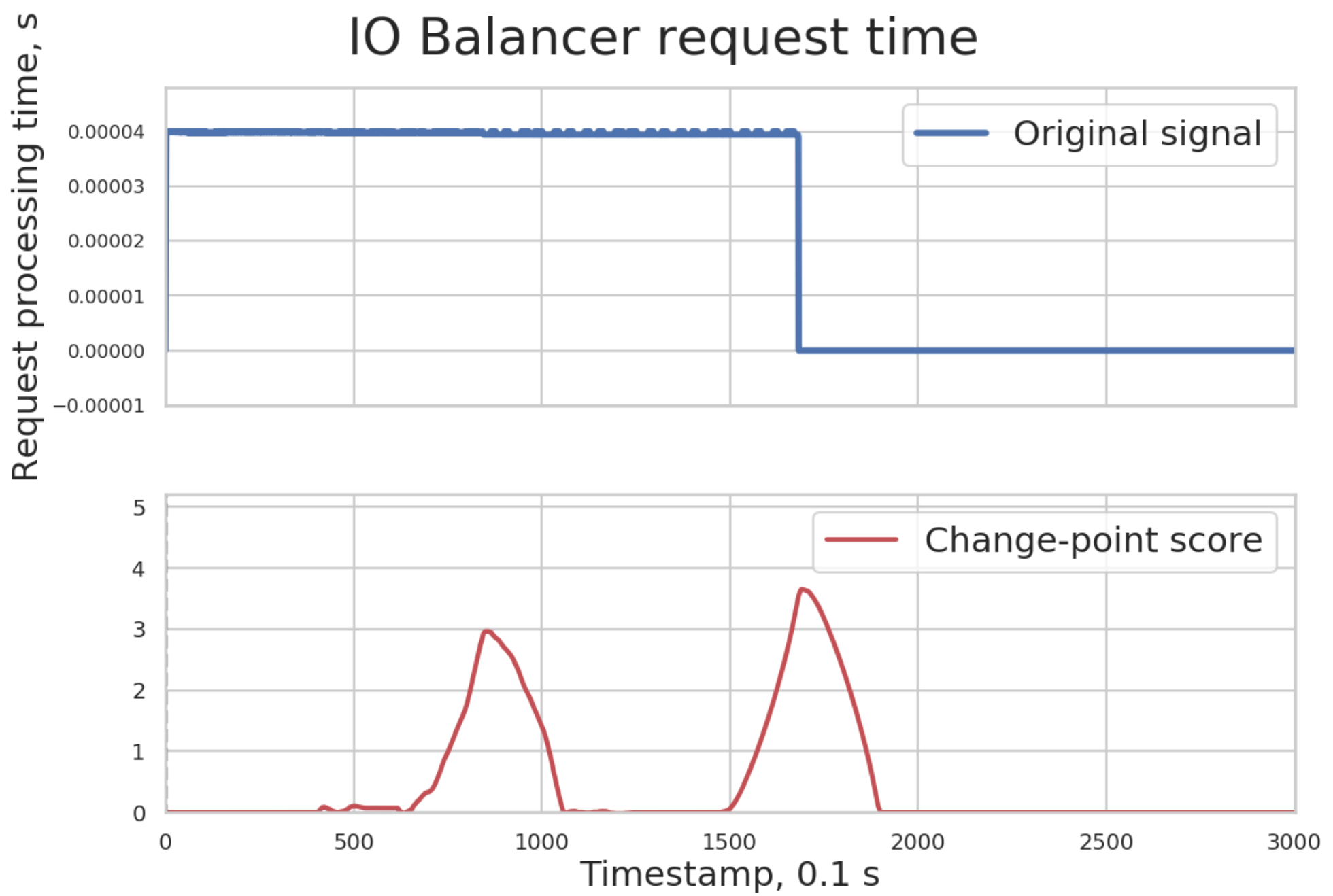}
\caption{\label{fig:cpd2} IO balancer time series}
\end{minipage} 
\end{figure}

\begin{figure}[h]
\begin{minipage}{19pc}
\includegraphics[width=19pc]{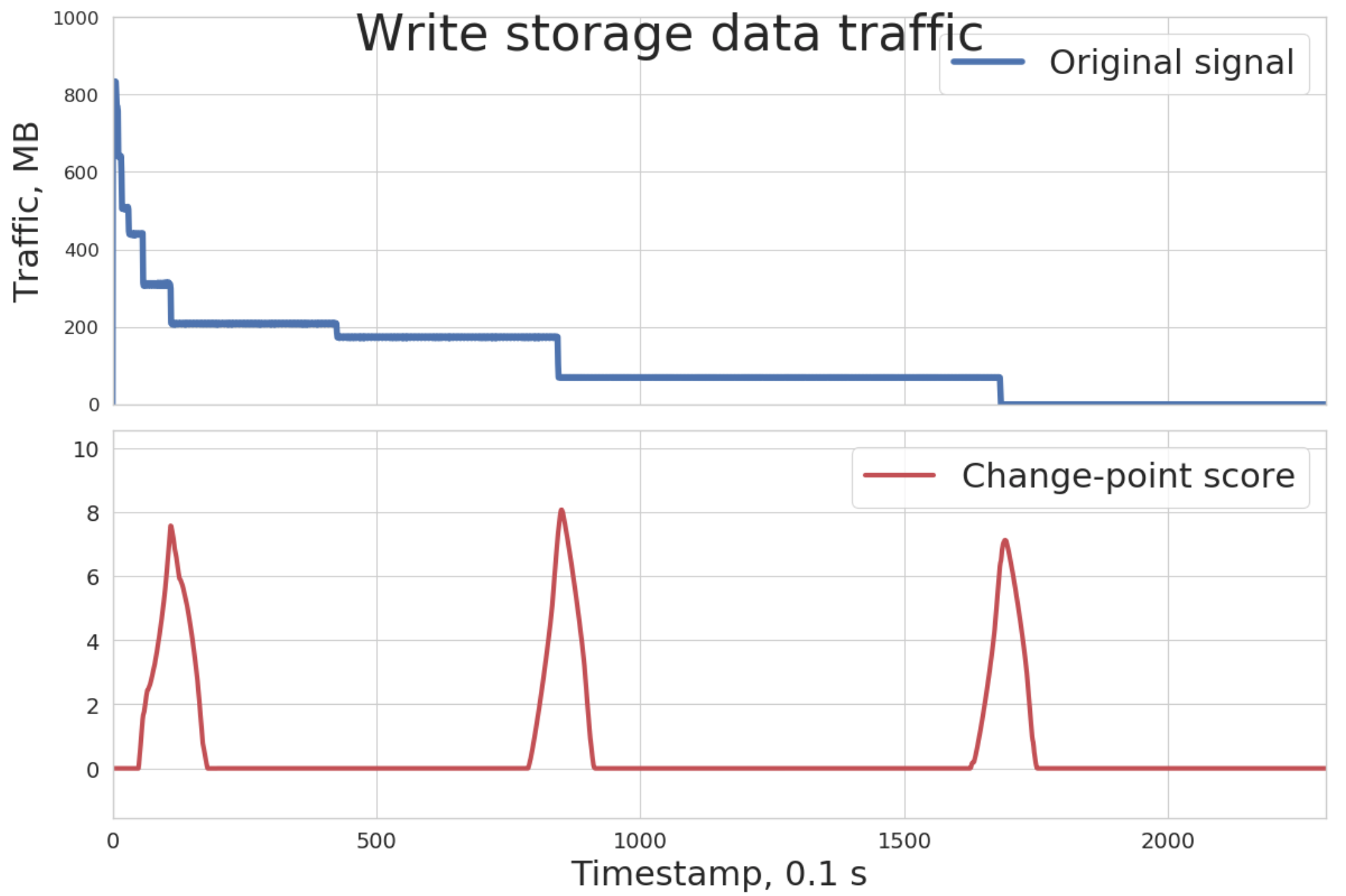}
\caption{\label{fig:cpd3}Storage traffic}
\end{minipage}\hspace{1pc}%
\begin{minipage}{19pc}
\includegraphics[width=19pc]{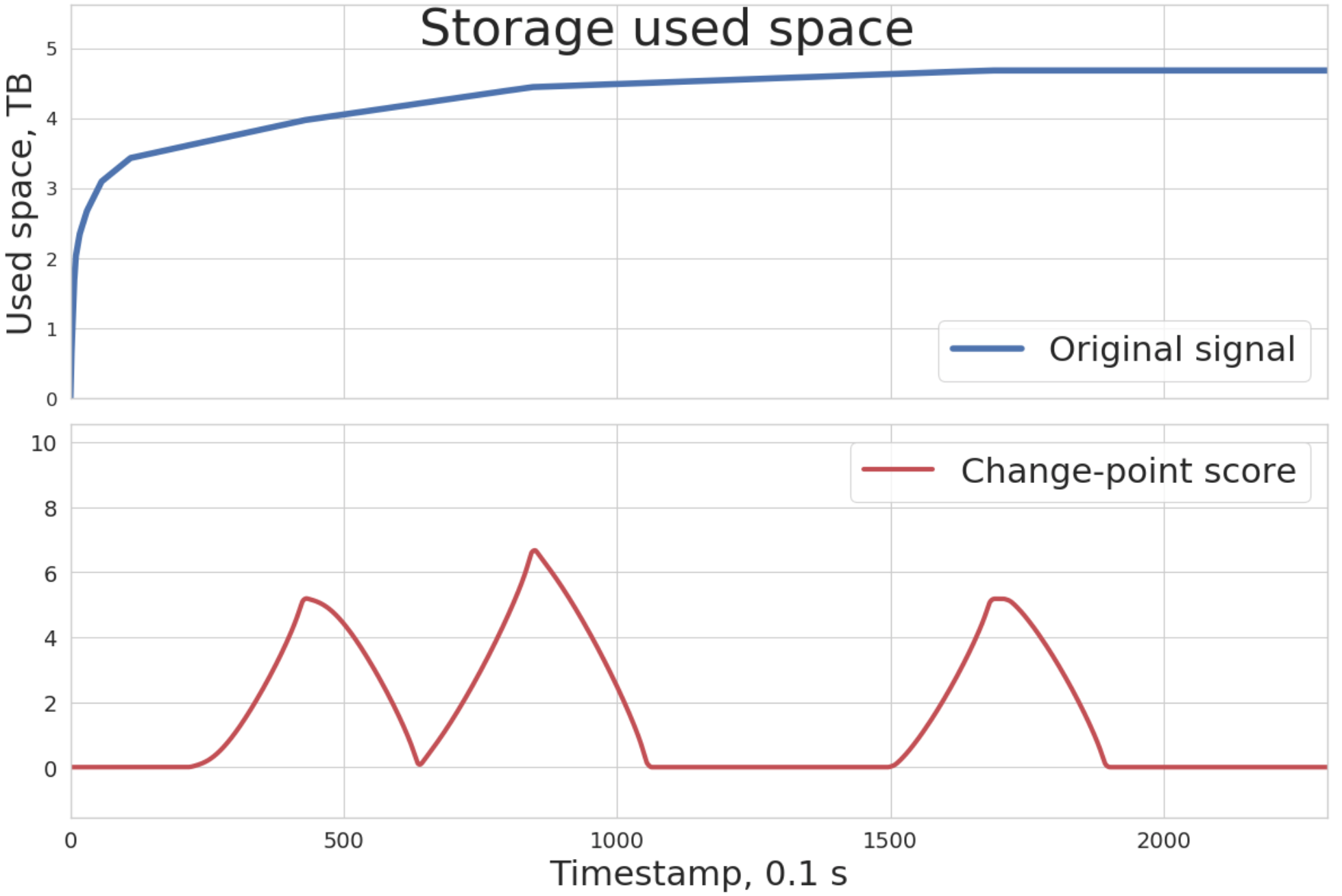}
\caption{\label{fig:cpd4}Storage used space}
\end{minipage} 
\end{figure}

To check the change-point algorithm against the simulation data, four time-series datasets were prepared: 1) controller's CPU load metric 2) load balancer request time 3) data traffic to storage devices and 4) differences change of used space. Their time-series are shown on the upper halves of Figures~\ref{fig:cpd1},~\ref{fig:cpd2},~\ref{fig:cpd3} and ~\ref{fig:cpd4}.

 As shown in the bottom halves of the figures above, the algorithm can identify data points where distribution changes. A red line on each plot is a CPD score line. The higher values it has, the more confident algorithm about a change point occurred at this timestamp. 

\section*{Conclusion}

The simulator for modeling storage infrastructure based on the event-driven paradigm was presented. It allows researchers to try different I/O load scenarios to test disk performance and model failures of its hardware components. By providing large amounts of synthetic data of anomalies and time series of a machine in various modes,  the simulator can also be used as a benchmark for comparing different change-point detection algorithms. In this work, the density ratio estimation CPD algorithm were successfully applied to the simulator data.

\ack{This  research  was  supported  in  part through  computational  resources  of  HPC  facilities  at NRU HSE.}

\section*{References}

\bibliography{ref}

\end{document}